\documentclass{article}

\PassOptionsToPackage{numbers, compress}{natbib}
\usepackage[final,main]{neurips_2026}

\usepackage[utf8]{inputenc} 
\usepackage[T1]{fontenc}    
\usepackage{hyperref}       
\usepackage{url}            
\usepackage{booktabs}       
\usepackage{amsfonts}       
\usepackage{nicefrac}       
\usepackage{microtype}      
\usepackage[table]{xcolor}         
\usepackage{amsmath}
\usepackage{graphicx}
\usepackage{subcaption} 
\usepackage{tabularx}
\usepackage{placeins}
\usepackage{floatflt}
\usepackage{algorithm}
\usepackage[capitalize]{cleveref}
\usepackage{graphicx}
\usepackage{tabularx}
\usepackage{wrapfig}
\usepackage{multirow}
\usepackage{enumitem}
\usepackage{xcolor}
\usepackage{amsmath}
\usepackage{booktabs}
\usepackage{xcolor}
\usepackage{pifont}
\usepackage{makecell}

\Crefname{equation}{Eq.}{Eqs.} 
\crefname{equation}{Eq.}{Eqs.} 

\definecolor{forth}{RGB}{255,255,255}  
\definecolor{third}{RGB}{255,240,192} 
\definecolor{second}{RGB}{192,216,255}   
\definecolor{best}{RGB}{255,192,192}
\definecolor{lightgray}{RGB}{180, 180, 180}

\title{Memorize When Needed: \\ Decoupled Memory Control for Spatially Consistent Long-Horizon Video Generation}

\usepackage{url}
\author{
{\bfseries Yanjun Guo$^{1,2,\star}$,
Zhengqiang Zhang$^{1,2,\star}$,
Pengfei Wang$^{1,\star}$,}\\
{\bfseries Xinyue Liang$^{1}$,
Zhiyuan Ma$^{1}$,
Lei Zhang$^{1,2,\dagger}$}\\
{$^{1}$The Hong Kong Polytechnic University \qquad $^{2}$OPPO Research Institute}\\
{\small yanjunn.guo@connect.polyu.hk \quad cslzhang@comp.polyu.edu.hk}\\
{\small $^{\star}$Equal contribution \qquad $^{\dagger}$Corresponding author}
}

\begin{document}
\maketitle
\renewcommand{\thefootnote}{${\dagger}$}
\footnotetext{This work is supported by the PolyU-OPPO Joint Innovative Research Center.}
\renewcommand{\thefootnote}{\arabic{footnote}}

\begin{abstract}

Spatially consistent long-horizon video generation aims to maintain temporal and spatial consistency along predefined camera trajectories. Existing methods mostly entangle memory modeling with video generation, leading to inconsistent content during scene revisits and diminished generative capacity when exploring novel regions, even when trained on extensive annotated data. To address these limitations, we propose a decoupled framework that separates memory conditioning from generation. Our approach significantly reduces training costs while simultaneously enhancing spatial consistency and preserving the generative capacity for novel scene exploration. Specifically, we employ a lightweight, independent memory branch to learn precise spatial consistency from historical observations.
We first introduce a hybrid memory representation to capture complementary temporal and spatial cues from generated frames, then leverage a per-frame cross-attention mechanism to ensure each frame is conditioned exclusively on the most spatially relevant historical information, which is injected into the generative model to ensure spatial consistency.
When generating new scenes, a camera-aware gating mechanism is proposed to mediate the interaction between memory and generation modules, enabling memory conditioning only when meaningful historical references exist.
Compared with the existing method, our method is highly data-efficient, yet the experiments demonstrate that our approach achieves state-of-the-art performance in terms of both visual quality and spatial consistency. Project page can be found at \url{https://iguoyanjun.github.io/memorize-when-needed-project-page/}.

\end{abstract}

\begin{figure*}[t]
	\centering
	\includegraphics[width=\linewidth]{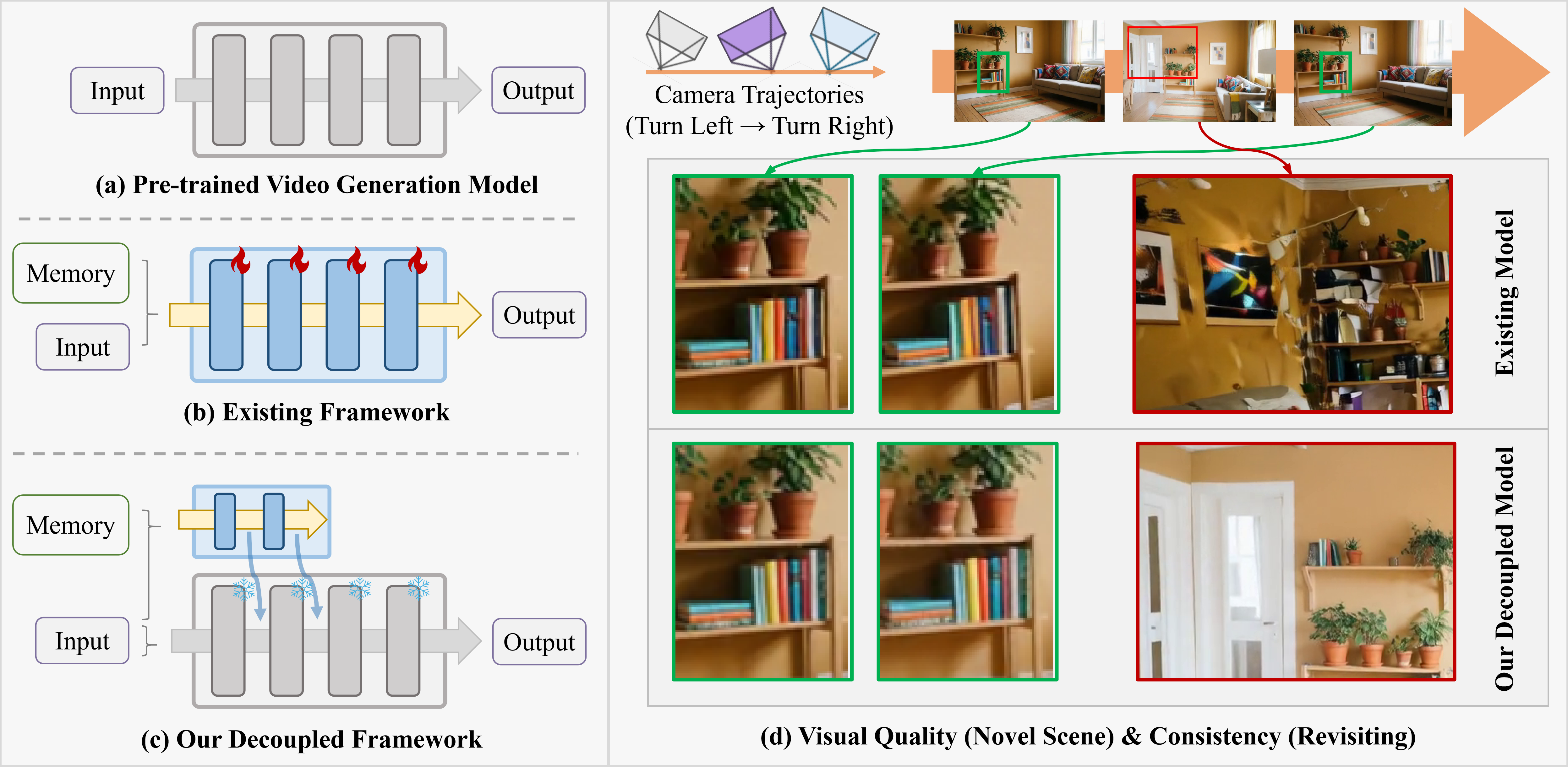}
	\caption{
\textbf{Comparison of model architecture of long-horizon video generation methods.} (a) A standard pre-trained video generation model. (b) Existing methods typically entangle generation and memory modeling within a unified fine-tuning framework. (c) In contrast, our model decouples memory modeling in a lightweight branch while keeping the pre-trained backbone frozen. (d) Our design enables realistic texture synthesis in novel scenes (e.g., the realistic door in second frame) while ensuring consistency in revisited locations (e.g., the consistent wooden shelf in the \textcolor{green}{green} box), outperforming existing methods that suffer from structural distortions (see \textcolor{red}{red} box).}
	\label{fig:fig1}
    \vspace{-3mm}
\end{figure*}

\vspace{-3mm}
\section{Introduction}
\label{sec:intro}

Recent state-of-the-art video generation models \cite{gao2025seedance,li2025hunyuan,bruce2024genie,kong2024hunyuanvideo,kling,wan2025wan,yang2024cogvideox} have achieved impressive spatio-temporal coherence within short-term sequences. However, extending such fidelity to long-horizon synthesis remains a fundamental challenge~\cite{duan2025worldscore,team2025hunyuanworld,kong2024hunyuanvideo}. This limitation is particularly obvious when scene revisit is required, where the model must maintain consistency with previously seen environments~\cite{bar2025navigation}. Due to the inherent memory bottleneck imposed by finite context windows, existing models~\cite{chen2024diffusion,song2025history,henschel2025streamingt2v}  often fail to perceive distant historical observations, ultimately leading to inconsistent content and visual discontinuities.

To overcome context limitations, many recent methods have introduced memory retrieval mechanisms~\cite{sun2025worldplay, yu2025context,xiao2025worldmem,li2025vmem}. By leveraging camera trajectories as geometric cues, these approaches dynamically fetch relevant historical frames to guide the synthesis, aiming to maintain visual consistency during revisiting.
For example, WorldMem~\cite{xiao2025worldmem} and Context-as-Memory~\cite{yu2025context} concurrently propose FOV-overlap scoring for memory selection, a practical strategy adopted by Hunyuanworld1.5~\cite{sun2025worldplay}.
Despite the potential of retrieval-augmented memory,
such an architecture entanglement introduces a compromise between memory adherence and generative quality, inevitably leading to suboptimal long-term consistency and degraded visual fidelity (e.g., the structural distortions shown in the \textcolor{red}{red} box of Fig.~\ref{fig:fig1}(d)), even after extensive fine-tuning.

We address these limitations with a decoupled memory framework that separates memory modeling from the generative process (see Fig.~\ref{fig:fig1}(c) vs. (b)). By freezing the video backbone and employing a lightweight auxiliary branch for memory injection, we eliminate architectural entanglement, preserving the rich priors in pretrained generation models and enabling consistent long-horizon synthesis without sacrificing the quality of novel scenes. As shown in Fig.~\ref{fig:fig1}(d), this design enables consistent long-horizon synthesis (e.g., the shelf) without sacrificing the visual fidelity of novel scenes (e.g., the door).
To adaptively modulate the memory injection, we introduce a camera-aware gating mechanism that activates memory conditioning only when informative references exist, effectively enabling the exploration of unseen regions. Our design balances the dual demands of long-horizon generation: enforcing spatial consistency during revisits while leveraging the backbone's generative priors for novel exploration. Moreover, this decoupled paradigm also reduces training overhead --- the memory branch can be optimized on generic videos using simple data augmentations, obviating the need for costly, manually annotated datasets.

In the memory control branch, we first retrieve relevant historical frames using a Field-of-View (FOV)-guided retrieving strategy, following~\cite{xiao2025worldmem}. We feed these frames into a video encoder to extract hybrid memory representations: continuous temporal tokens capture motion patterns and scene dynamics, while fine-grained spatial tokens provide sharp, static references for the current viewpoint. This combination yields complementary contextual and geometric cues.
Subsequently, to achieve frame-level visual alignment, we employ per-frame cross-attention blocks. These blocks allow the generative tokens to selectively query only the most spatially aligned historical frame from the hybrid representation via attention masking. Our design serves a dual purpose: ensuring precise control by filtering out irrelevant spatial noise, and reducing computational overhead by avoiding redundant attention across the entire history buffer. Finally, the aligned memory features are injected into the frozen backbone via the aforementioned camera-aware gating mechanism. Our experiments demonstrate the effectiveness of our decoupled memory control design, which is not only much more data-efficient than existing methods, but also achieves state-of-the-art video quality and spatial consistency, especially for scene revisits.  

\vspace{+1mm}
In summary, our main contributions are as follows:
\vspace{-2mm}
\begin{itemize}
    \item We propose a decoupled memory control method with a camera-aware gating mechanism. This design effectively balances the preservation of pretrained generative priors for novel scene exploration and the effective memory injection for long-horizon video consistency.   
    \item We extract hybrid memory representations from historical frames and employ a masked per-frame cross-attention module to achieve precise frame-level alignment. This approach leverages both motion patterns and spatial details for accurate context retrieval.

    \item Experiments demonstrate that our method achieves state-of-the-art performance in terms of visual quality and spatial consistency, especially for revisited scenes. Notably, our method is highly data-efficient, capable of learning robust memory control from generic videos using simple augmentations.
    
\end{itemize}
\section{Related Work}

\textbf{Video Generation and Controllability.}
Video generation has evolved from early UNet-based diffusion models~\cite{chen2024videocrafter2,blattmann2023stable} to large-scale latent diffusion transformers~\cite{veo,gao2025seedance,li2025hunyuan,bruce2024genie,kong2024hunyuanvideo,kling,hailuo,wan2025wan,yang2024cogvideox}.
Benefiting from advances in model architectures~\cite{ho2020denoising,peebles2023scalable,rombach2022high} and large-scale video data curation pipelines \cite{li2025hunyuan,wan2025wan}, foundation text/image-to-video models exhibit zero-shot generalization with consistent physical dynamics and notable 3D consistency over short temporal ranges \cite{wiedemer2025video}.
Recent studies have also investigated whether video models can reliably capture spatio-temporal structure beyond perceptual realism, exploring verifiable spatio-temporal reasoning~\cite{Zhu2026VideoModelsReason} and robust multimodal grounding~\cite{Wen2026VisionSpeaksSound}.
To enable viewpoint control, recent efforts inject camera poses into pretrained video diffusion models via conditioning modules, using camera parameters~\cite{wang2024motionctrl}, dense Plücker ray embeddings~\cite{bahmani2025ac3d,he2024cameractrl}, or learnable pose tokens~\cite{guo2023animatediff,li2025cameras,miyato2023gta,su2024roformer,sun2024dimensionx}; other approaches use synthetic game-engine data with discrete action commands~\cite{bar2025navigation,oasis2024,he2025matrix,sun2025virtual,valevski2024diffusion,yu2025gamefactory} or enforce 3D consistency through depth- and pose-based frame warping~\cite{cao2025uni3c,ren2025gen3c,yu2025trajectorycrafter,yu2024viewcrafter}.


\noindent\textbf{Long-Term Context and Memory Modeling.}
Generating long videos that extend beyond the native temporal window of pretrained models is a challenging issue, given the quadratic attention complexity. Training-free methods~\cite{lu2024freelong,zhao2025riflex,xi2025sparse} extend pretrained models through noise rescheduling, temporal-frequency rebalancing, or sparse attention. Diffusion Forcing~\cite{chen2024diffusion} and History-Guidance~\cite{song2025history} condition denoising on previous frames with decayed noise, and are scaled by SkyReels-V2~\cite{chen2025skyreels} and Magi-1~\cite{teng2025magi}. {Meanwhile, recent multi-shot video generation methods~\cite{An2026OneStory} explicitly model cross-shot dependencies through fixed-window attention or keyframe conditioning to improve narrative consistency across shots.} Other works distill bidirectional diffusion into causal generators~\cite{cui2025self, huang2025self,kim2024fifo,yang2025longlive} to reduce autoregressive error accumulation, while ~\cite{henschel2025streamingt2v,zhang2025frame} augment pretrained models with memory modules and generate long videos iteratively, achieving high visual quality.  

Beyond temporal coherence, long-horizon video generation also requires maintaining spatial consistency when revisiting previously observed regions. Existing methods use either explicit 3D reconstruction~\cite{cao2025uni3c,ren2025gen3c,yu2025trajectorycrafter,yu2024viewcrafter,huang2025memory,wu2026infiniteworld,wang2026one2scene,An2026VGGRPO} or implicit context conditioning~\cite{li2025vmem,xiao2025worldmem,yu2025context,hong2025relic,sun2025worldplay}. Explicit methods provide strong geometric guidance but heavily depend on reconstruction quality, introducing additional computational overhead. Implicit methods retrieve spatially relevant history from memory banks, e.g., through Field-of-View overlap scoring~\cite{xiao2025worldmem,yu2025context}, offering a practical solution for maintaining spatial coherence. 
However, these methods entangle memory processing with the generation backbone (e.g., via input concatenation or modified attention), leading to high training costs, inflexible conditioning, and over-reliance on history that reduces diversity in novel regions.
\section{Method}

In this section, we first formulate the problem of long-horizon consistent video generation. We then present our decoupled framework, detailing the camera-aware gating mechanism, the architectures of the memory control branch (including hybrid memory representation and per-frame cross-attention blocks), and the tailored training strategies designed to handle complex exploration trajectories.

\subsection{Preliminaries}
\renewcommand{\floatpagefraction}{0.9}

Long-horizon video generation aims to synthesize video sequences that maintain long-term spatial consistency along a predefined camera trajectory. Given an initial image $I_0$, a text prompt $\mathcal{T}$, and a sequence of camera poses $\{P_t\}_{t=1}^{T}$, the goal is to generate a sequence of frames $\{I_t\}_{t=1}^{T}$ that follow the specified viewpoint changes while preserving scene consistency.
In latent diffusion-based I2V models, the initial image $I_0$ conditions DiT-based iterative denoising of video latents, which are finally decoded into video sequences by a 3D VAE.

A core challenge is how to maintain spatial consistency, particularly when the camera revisits previously observed locations.
Prior works~\cite{yu2025context,li2025vmem,he2025cameractrl} typically adopt a segment-based iterative approach, dividing the sequence into $K$ consecutive segments $\{S_k\}_{k=1}^{K}$. In this pipeline, each segment is generated by utilizing previously generated content as conditioning context:
\begin{equation}
	S_k = V_\theta(I_{\text{pre}}, \mathcal{P}_{S_k}, \mathcal{M}_k, \mathcal{T}), \quad \{I_t\}_{t=1}^{T} = \text{Concat}(\mathcal{D}(S_1), \dots, \mathcal{D}(S_K)),
\end{equation}
where $I_{\text{pre}}$ denotes the last frame of the segment $S_{k-1}$ ( $S_0 = \{I_0\}$), $\mathcal{P}_{S_k}$ is the set of camera poses corresponding to segment $S_k$, and $\mathcal{M}_k = \{(I_0, (\mathcal{D}(S_1), P_1), \ldots, \\ (\mathcal{D}(S_{k-1}), P_k)\}$ represents the accumulated visual memory and camera poses from the initial image and all preceding segments. $\mathcal{D}$ is the video decoder.

\begin{figure*}[t]
	\centering
	\includegraphics[width=\linewidth]{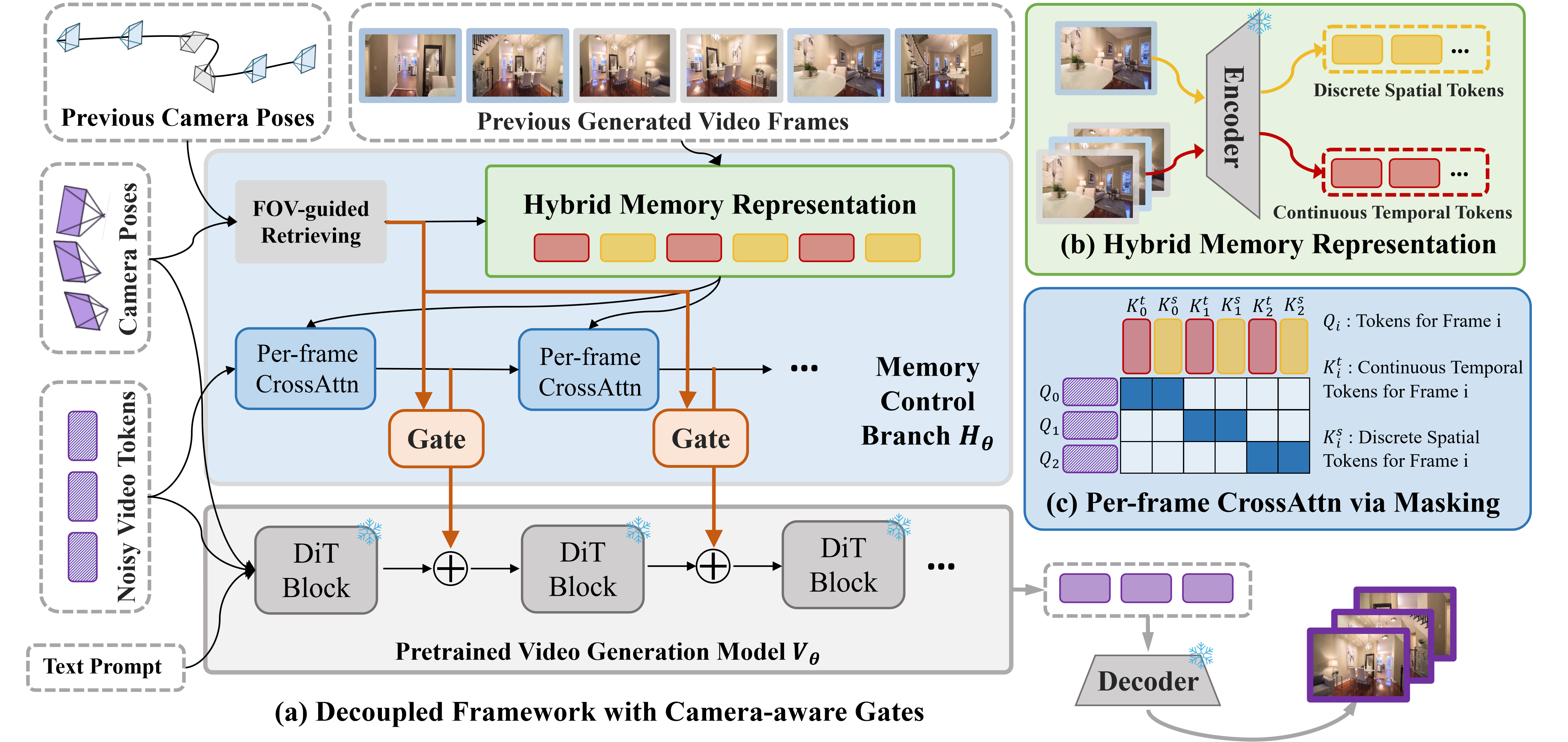}
\caption{\textbf{Overview of our decoupled framework.} (a) We disentangle memory control (\textcolor{blue}{blue} branch) from the generative backbone (\textcolor{gray}{gray} branch), employing a camera-aware gating mechanism (\textcolor{orange}{orange} part) to adaptively modulate memory injection. (b) The hybrid memory representation (\textcolor{green}{green} part) extracts complementary spatio-temporal cues from history frames. (c) Per-frame cross-attention blocks (\textcolor{blue}{blue} part) enforce frame-level alignment, where each latent token attends solely to its spatially corresponding historical frame, ensuring both precise control and computational efficiency.}
	\label{fig:network}
    \vspace{-3mm}
\end{figure*}

\subsection{Motivation and Framework Overview}
\label{sec:framework}

Existing methods generally entangle memory modeling with video generation within a unified framework. The model $V_\theta$ is {burdened with the dual task of} creating new content and enforcing consistency constraints simultaneously. This often leads to a trade-off, compromising either the perceptual quality of generated frames or the long-term spatial consistency. 
To address this limitation, we propose to decouple these objectives into separate streams. Specifically, we retain $V_\theta$ for high-fidelity video generation while introducing an additional lightweight memory control branch $H_\theta$ dedicated to processing historical memory. The generation process is thus reformulated as:
\begin{equation}
	S_k = V_\theta(I_{\text{pre}}, \mathcal{P}_{S_k}, H_\theta(\mathcal{M}_k, \mathcal{P}_{S_k}), \mathcal{T}),
\end{equation}
where $H_\theta(\cdot)$ extracts compact, consistency-aware features from the raw memory $\mathcal{M}_k$, allowing $V_\theta$ to focus primarily on synthesizing novel content aligned with the text prompt and camera trajectory.

Our architecture is illustrated in Fig.~\ref{fig:network}(a), which consists of a frozen {video generation backbone $V_\theta$} (the \textcolor{gray}{gray} branch) dedicated to high-quality synthesis, and a lightweight {memory control branch $H_\theta$} (the \textcolor{blue}{blue} branch) designed to extract relevant historical context. 
In this decoupled framework, the inherent conflict between maintaining long-term consistency and generating diverse novel content is resolved by isolating their respective optimization goals. The backbone focuses solely on visual fidelity and are free from the constraints of historical alignment, while the memory branch specializes in context extraction. 

\subsection{Camera-aware Gating Mechanism}
\label{sec:gating}
The core design principle of our framework is to inject historical information \textit{only when needed}.Indiscriminately enforcing memory constraints can impede novel content generation in previously unseen regions and introduce artifacts in regions with unreliable memory retrieval. To strike a balance between consistency and creativity, we introduce an on-the-fly {camera-aware gating mechanism} (the \textcolor{orange}{orange} part in Fig.~\ref{fig:network}(a)) that dynamically modulates the influence of the memory branch based on the exploration trajectory.
Unlike previous works that must interact with $\mathcal{M}_k$ in a segment-wise manner, our approach regulates the interaction between the two streams ($V_\theta$ and $H_\theta$) in a frame-wise manner via a gating mechanism. This provides more precise control for video generation while minimizing unintended artifacts. 

Specifically, given camera poses $P_{S_k}$ for the current segment and the corresponding history $\mathcal{M}_k$, we first obtain the geometric relevance scores $s_i$ that incorporate both FOV overlap
and camera distance between each current pose $P_i \in P_{S_k}$ and all historical poses. 
The geometric score $s_i$ provides the initial gating decision, which is further refined by translation-distance and temporal-redundancy filtering, yielding the final binary gate $g_i \in \{0,1\}$ for each frame. Details are provided in Appendix.
Let $\mathbf{f}^i_l$ denote the intermediate features of $V_\theta$ at layer $l$ for frame $I_i$, and $\mathbf{h}^i_l$ represent the corresponding features from $H_\theta$. The feature injection is formulated as:
\begin{equation}
    \mathbf{f}^i_l = \mathbf{f}^i_l + g_i \cdot \mathbf{h}^i_l,
\end{equation}
where $g_i$ serves as a \textbf{hard gate}: it enables memory injection only for frames with reliable historical correspondences, which act as anchors for the segment. For novel or unreliably matched frames, the memory connection is disabled, allowing the backbone to generate content without  unnecessary or erroneous memory constraints while maintaining consistency through segment-level temporal coherence. This mechanism effectively harmonizes generative capability with consistency constraints.

\subsection{Memory Control Branch}
\label{sec:memory_rep}

As illustrated in Fig.~\ref{fig:network}(a), our memory control branch first identifies relevant historical frames and segments via an FOV-guided retrieval strategy (similar to that in~\cite{xiao2025worldmem}). Subsequently, it constructs a hybrid memory representation through per-frame cross-attention blocks to aggregate pertinent information. The details are presented as follows.

\textbf{Hybrid Memory Representation.}
The efficacy of the memory control branch hinges on the quality and reliability of features retrieved from historical frames. 
To enhance robustness to the imperfect matches introduced by FOV-based retrieval, as depicted in Fig.~\ref{fig:network}(b), we propose a hybrid memory representation that combines complementary chunk-level and frame-level cues from the retrieved history.
One is continuous temporal memory, which encodes a short window of consecutive historical frames to aggregate multi-frame evidence $K^{t}$ around the retrieved frames, thereby improving robustness to single-frame retrieval errors.
The other is discrete spatial memory to provide a sharp reference for the current viewpoint. We obtain those discrete spatial tokens $K^{s}$ by inputting individual retrieved frames.
Together, the chunk-level memory offers stable multi-frame context, while the frame-level memory preserves precise appearance details.

\textbf{Per-frame Cross-Attention Blocks.}
To extract relevant information for current latents, we employ them as queries to retrieve useful context from historical tokens ($K^{s}$ and $K^{t}$) within cross-attention blocks, where camera poses of queries and keys are injected as additional positional embeddings.
To precisely control the interaction between those historical tokens and current video latents, we apply a mask strategy to impose a strict constraint where each latent interacts only with the tokens from its corresponding history frame, as shown in Fig.~\ref{fig:network} (c). This localized attention mechanism not only reduces computational overhead but also alleviates optimization difficulties.

\subsection{Training Strategies}
\label{sec:training}

A key advantage of our decoupled architecture is that it streamlines the optimization of the memory control branch. Specifically, it allows us to augment existing real-world videos to construct revisiting and exploration data, circumventing the need for expensive 3D dataset rendering. We synthesize training sequences by strategically sampling and reordering video frames.

\textbf{Synthesizing Pseudo-Loops for Memory Training.} To train the model on revisiting events using standard videos, we synthesize sequences by reorganizing frame orders to simulate closed-loop trajectories (e.g., creating forward-backward loops). A naive approach --- using the same frame as both history and target --- would allow the model to learn a trivial identity mapping. To avoid this, we apply a temporal stride strategy: we use a frame at time $I_t$ as the history reference but require the model to generate its neighbor at $I_{t+\delta}$. This controlled perturbation between the memory reference and the target view mimics the imperfect matches produced by FOV-based retrieval and encourages the memory branch to learn robust content-level correspondence from geometrical cues.


\textbf{History Dropout for Novel Scene Exploration.} To ensure the model can seamlessly transition between revisiting known regions and exploring new ones, we introduce a history dropout strategy. During training, we randomly mask out the historical reference frames for certain segments. This compels the model to dynamically adapt: when history is unavailable (masked), it must rely on the backbone's generative priors to synthesize novel content; when history is present, it utilizes the memory for consistency. This simple regularization effectively guides the model to activate memory only when valid references exist.

\section{Experiment}
\label{sec:experiment}
This section presents extensive experiments to validate our method.
We begin by outlining the experimental setup and protocols in Sec.~\ref{sec:exp_setup}, followed by a comparative analysis against existing methods in Sec.~\ref{sec:main_results}.
Sec.~\ref{sec:additional_exp} further evaluates the robustness of our method on out-of-distribution data and challenging trajectory patterns.
Ablation studies are provided  in Sec.~\ref{sec:ablation_studies}, where we analyze the effects of the hybrid memory representation and camera-aware gating mechanism.

\subsection{Experimental Setup}
\label{sec:exp_setup}

\noindent\textbf{Training Settings.}
We adopt two representative image-to-video (I2V) diffusion models as our generative backbones: Wan2.1-14B-I2V~\cite{wan2025wan} for high-quality generation and CogVideoX-5B-I2V~\cite{yang2024cogvideox} for practical efficiency.
Following AC3D~\cite{bahmani2025ac3d}, we utilize Pl\"ucker coordinates for camera pose representation and incorporate an additional module for precise camera control.
During memory branch training, we freeze the parameters of both the video backbone and the camera control module, optimizing only the separate memory control branch.
We use the Adam optimizer with a learning rate of $1\times 10^{-4}$.
The effective batch size is set to 16, and the models are trained on NVIDIA A800 GPUs for 10K iterations. More training details are provided in the \textbf{Appendix}.

\vspace{+1mm}
\noindent\textbf{Training and Testing Data.}
For training, we utilize the RealEstate10K~\cite{Stereo} dataset to synthesize video sequences that simulate revisiting and exploration scenarios (details in Sec.~\ref{sec:training}). The data is processed into 49-frame video segments at 480p resolution.
For testing, following the protocol in WorldPlay~\cite{sun2025worldplay} and VMem~\cite{li2025vmem}, we sample trajectories from RealEstate10K and extend them with a return path along the initial trajectory for consistency evaluation.

\label{sec:challenging_traj}
{Furthermore, to assess generalization, we collect 200 additional test scenes sourced from both the internet and AI models, which cover indoor and outdoor environments and span diverse content and visual styles.}
We also design a suite of challenging trajectory patterns to rigorously evaluate consistency under diverse conditions. (a) \textit{Repeated Revisiting}: where the camera rotates forth and back to the same viewpoint multiple times, creating a repetitive pattern that tests the model's stability against error accumulation; (b) \textit{Random Loop Insertion}: introducing return loops of varying lengths at random positions along a trajectory; and (c) \textit{Spatially-Offset Return}: where the camera returns via a slightly shifted path instead of exact retracing, testing robustness to viewpoint deviations.

\noindent\textbf{Evaluation Metrics.}
We evaluate our method from two primary perspectives: long-term consistency in revisiting scenarios and generation quality on novel scenes.
For \textit{long-term consistency}, following~\cite{li2025vmem,sun2025worldplay,yu2025context}, we employ PSNR, SSIM, LPIPS as quantitative metrics. Specifically, we utilize cyclic camera trajectories that revisit previously observed viewpoints, comparing each generated frame on the return path with its counterpart from the initial pass. Higher PSNR/SSIM scores and lower LPIPS scores indicate superior consistency.
For \textit{generation quality}, we assess the synthesized videos across both spatial and temporal dimensions, following VBench~\cite{huang2024vbench}.
We use Aesthetic Quality (AQ)\cite{schuhmann2022laion} and Image Quality (IQ)\cite{ke2021musiq} to measure image quality, and adopt Motion Smoothness (MS)\cite{li2023amt} and Dynamic Degree (DD)\cite{teed2020raft} to estimate motion magnitude.

\vspace{+1mm}
\noindent\textbf{Compared Methods.}
We conduct comprehensive comparisons with existing methods, categorized into two groups.
(1) \textit{Models without memory}: AC3D\cite{bahmani2025ac3d} (re-implemented on CogVideoX-5B-I2V), DFoT\cite{song2025history}, and SEVA\cite{zhou2025stable}. These methods support camera-controlled generation, but lack explicit mechanisms for maintaining cross-clip consistency.
(2) \textit{Models with memory}: VMem\cite{li2025vmem} and WorldPlay~\cite{sun2025worldplay}. For WorldPlay, we evaluate both its bidirectional attention variant (WorldPlay) and the final distilled version (WorldPlay-d).

\definecolor{first}{RGB}{255, 0, 0}   
\definecolor{second}{RGB}{0, 0, 255} 

\begin{table*}[t]
\centering
\caption{Long-term consistency and video quality comparisons on RealEstate10K. \textcolor{red}{\textbf{Red}}: best, \textcolor{blue}{blue}: second best in each column.}
\label{tab:main_results}
\resizebox{\textwidth}{!}{%
\setlength{\tabcolsep}{1pt}  
\begin{tabular}{l ccc cccc cccc}
\toprule
\multirow{3}{*}{\makecell{Method}} & \multicolumn{7}{c}{\textbf{Revisiting}} & \multicolumn{4}{c}{\textbf{Exploration}} \\
\cmidrule(lr){2-8} \cmidrule(lr){9-12}
& \multicolumn{3}{c}{{long-term consistency}} & \multicolumn{4}{c}{{video quality}} & \multicolumn{4}{c}{{video quality}} \\
\cmidrule(lr){2-4} \cmidrule(lr){5-8} \cmidrule(lr){9-12}
& \makecell{PSNR$\uparrow$} & \makecell{SSIM$\uparrow$} & \makecell{LPIPS$\downarrow$} & \makecell{AQ(\%)$\uparrow$} & \makecell{IQ(\%)$\uparrow$} & \makecell{MS(\%)$\uparrow$} & \makecell{DD(\%)$\uparrow$} & \makecell{AQ(\%)$\uparrow$} & \makecell{IQ(\%)$\uparrow$} & \makecell{MS(\%)$\uparrow$} & \makecell{DD(\%)$\uparrow$} \\
\midrule
AC3D\cite{bahmani2025ac3d}       
& 13.13 & 0.52 & 0.52 
& \textcolor{second}{51.37} & 64.71 & \textcolor{second}{99.29} & \textcolor{second}{95.24} 
& \textcolor{first}{\textbf{49.88}} & 63.84 & 99.21 & 92.86 \\
DFoT\cite{song2025history}               
& 14.93 & 0.48 & 0.34 
& 43.00 & 66.80 & 98.58 & 35.71 
& 42.20 & 65.71 & 98.67 & 35.71 \\
SEVA\cite{zhou2025stable}            
& 14.80 & 0.60 & 0.47 
& 42.78 & 61.67 & 98.40 & 88.10 
& 44.44 & 64.62 & 98.28 & 88.10 \\
VMem\cite{li2025vmem}    
& 15.37 & 0.65 & 0.37 
& 42.34 & 59.80 & 97.66 & 78.57 
& 42.38 & 59.37 & 98.04 & 80.95 \\
WorldPlay\cite{sun2025worldplay} 
& 16.31 & 0.64 & 0.36 
& \textcolor{first}{\textbf{51.51}} & 70.45 & 99.06 & 92.86 
& 49.24 & \textcolor{second}{70.57} & 99.04 & 92.86 \\
WorldPlay-d\cite{sun2025worldplay} 
& 15.09 & 0.63 & 0.31 
& 51.34 & \textcolor{second}{71.43} & \textcolor{first}{\textbf{99.32}} & 88.10 
& 49.00 & 70.12 & \textcolor{second}{99.32} & 90.48 \\
\midrule
Ours-Cog                      
& \textcolor{second}{20.23} & \textcolor{second}{0.69} & \textcolor{second}{0.23} 
& 50.64 & 63.53 & 99.17 & 92.86 
& 49.20 & 63.49 & \textcolor{first}{\textbf{99.42}} & \textcolor{second}{95.24} \\
Ours-Wan                      
& \textcolor{first}{\textbf{21.85}} & \textcolor{first}{\textbf{0.71}} & \textcolor{first}{\textbf{0.20}} 
& 51.24 & \textcolor{first}{\textbf{72.27}} & 98.86 & \textcolor{first}{\textbf{97.62}} 
& \textcolor{second}{49.55} & \textcolor{first}{\textbf{72.48}} & 98.88 & \textcolor{first}{\textbf{95.56}} \\
\bottomrule
\end{tabular}
}
\end{table*}
\subsection{Main Results}
\label{sec:main_results}

\begin{figure*}[t]
	\centering
	\includegraphics[width=\linewidth]{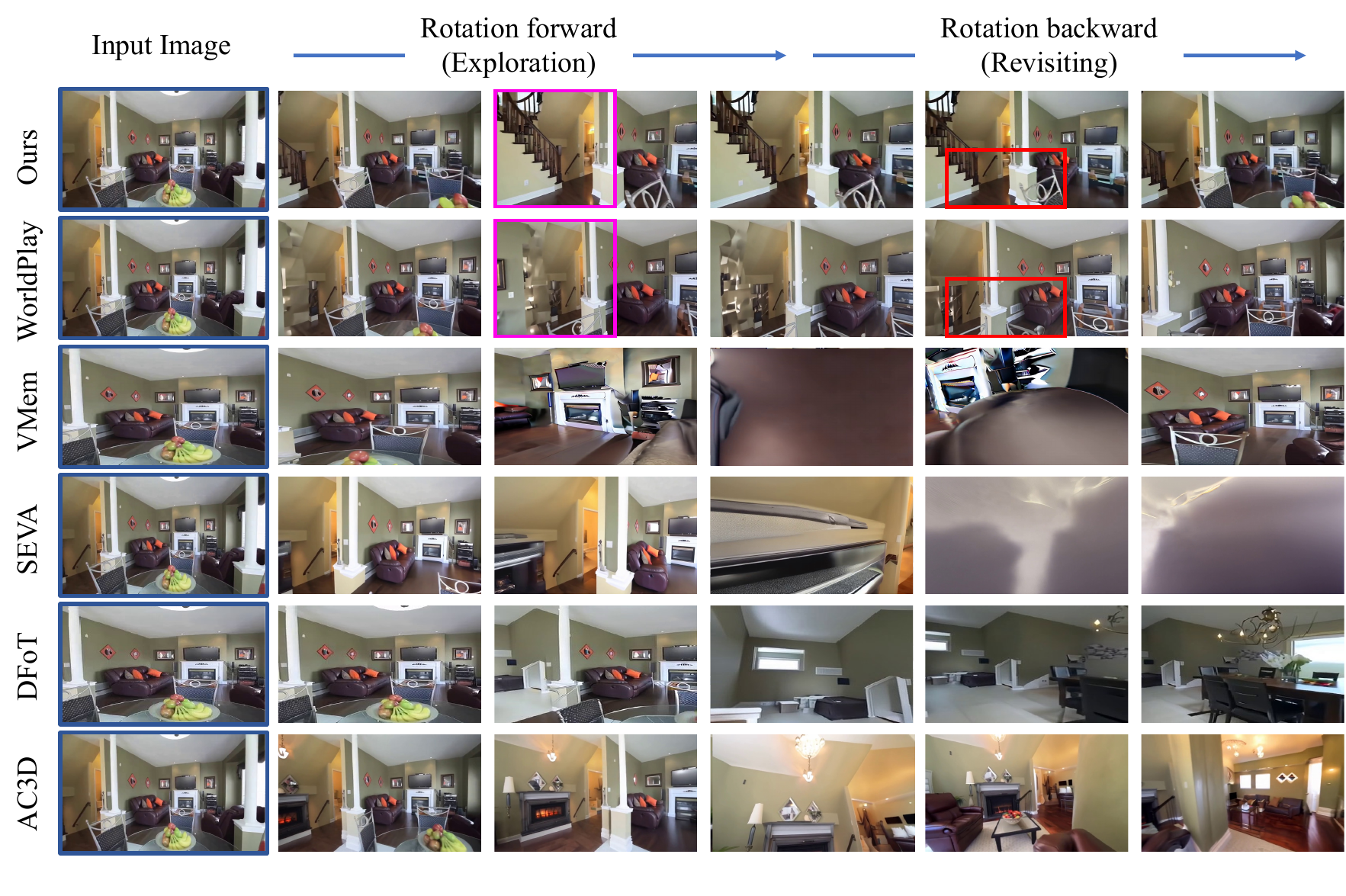}
    \vspace{-5mm}
\caption{Visual comparison. Our method generates clearly structured staircases in unseen regions (\textcolor{magenta}{pink} boxes) and faithfully reproduces fine-grained details such as the two chairs when revisiting the original scene (\textcolor{red}{red} boxes).} 
	\label{fig:experiment}
\vspace{-5mm}
\end{figure*}

\noindent\textbf{Quantitative Results.}
As shown in Table~\ref{tab:main_results}, our method (Ours-Wan) achieves state-of-the-art performance in both consistency and generation quality metrics.
Methods without explicit memory modeling --- such as AC3D\cite{bahmani2025ac3d}, DFoT\cite{song2025history}, and  SEVA\cite{zhou2025stable} --- struggle to maintain consistency over extended trajectories, resulting in inferior PSNR, SSIM, and LPIPS scores in the \textit{Revisiting} phase. Incorporating memory modeling notably improves consistency. In specific, WorldPlay~\cite{sun2025worldplay} achieves improved values of 16.31 (PSNR), 0.64 (SSIM), and 0.36 (LPIPS). Our decoupled framework surpasses these competitors by a significant margin, achieving superior results of 21.85 (PSNR), 0.71 (SSIM), and 0.20 (LPIPS), respectively. This demonstrates that our separate memory control branch provides significantly more precise memory retrieval and faithful scene reproduction.
Regarding video quality, while WorldPlay achieves a marginally higher AQ (51.51\%) compared to ours (51.24\%), our method outperforms all competitors in both Image Quality (IQ, 72.27\%) and Dynamic Degree (DD, 97.62\%).
This indicates that our approach effectively preserves high-fidelity generative capabilities and temporal coherence even in complex revisiting scenarios, without sacrificing visual details.

We also evaluate all methods on the \textit{Exploration} columns in Table~\ref{tab:main_results} to assess their generative capacity on novel scenes.
Our method (CogVideoX-based) achieves AQ on par with AC3D \cite{bahmani2025ac3d}, demonstrating that the decoupled design preserves the generative prior of the backbone and ensures stable generation capability during exploration.
Ours-Wan achieves the highest IQ (72.48\%) and DD (95.56\%) in the exploration phase with only a slight performance drop in AQ (1.69 points) compared to WorldPlay.
This demonstrates the effectiveness of our dynamic gating mechanism, which successfully balances retrieving memory for consistency and generating novel content for exploration.

\noindent\textbf{Qualitative Results.}
We provide qualitative comparisons in Figure~\ref{fig:experiment}, where the camera first rotates forward into unseen regions and then moves back to the starting viewpoint.  Methods AC3D~\cite{bahmani2025ac3d}, DFoT~\cite{song2025history} and SEVA~\cite{zhou2025stable} lack explicit cross-clip spatial memory, hallucinating mismatched content when the camera rotates back. Although VMem~\cite{li2025vmem} incorporates memory through image-conditioned generation, it suffers from severe error accumulation over extended sequences, synthesizing unreasonable leather texture during exploration. WorldPlay~\cite{sun2025worldplay} produces noticeable artifacts when the camera moves into unseen regions. In contrast, our method synthesizes coherent novel content, such as clearly structured staircases and a bathroom with an open door. When the camera rotates back, our method faithfully reproduces the original living room scene, including fine-grained details such as the two chairs beside the dining table, demonstrating the capability of our decoupled framework in both novel scene exploration and revisited scene reconstruction.

\begin{table*}[t]
\centering
\caption{Comparison of training data, parameters, and computational cost.}
\label{tab:efficiency}
\resizebox{0.85\textwidth}{!}{%
\begin{tabular}{l c c c c c c c}
\toprule
Method 
& \multirow{2}{*}{\shortstack{Training\\Data}} 
& \multicolumn{3}{c}{Parameters} 
& \multicolumn{3}{c}{FLOPs (TFLOPs)} \\
\cmidrule(lr){3-5} \cmidrule(lr){6-8}
&  
& Backbone
& Memory
& Total
& Backbone
& Memory
& Total \\
\midrule
VMem\cite{li2025vmem}
& - 
& 1.26B
& 0 
& 1.26B
& 67.46 
& 44.96
& 112.44  \\
WorldPlay~\cite{sun2025worldplay}
& 320K 
& 8.56B 
& 0 
& 8.56B
& 923.03 
& 3065.00 
& 3988.03 \\
Ours-Cog
& 14K 
& 5.60B 
& 128.42M 
& 5.73B
& 440.81 
& 2.97 
& 443.78 \\
Ours-Wan 
& 14K & 14.75B & 1371.38M & 16.12B
& 1240.22 & 71.62 & 1311.84 \\
\bottomrule
\end{tabular}
}
\vspace{-2mm}
\end{table*}

\definecolor{first}{RGB}{255, 0, 0}   
\definecolor{second}{RGB}{0, 0, 255} 

\begin{table*}[t]
\centering
\caption{Out-of-distribution quantitative comparison. \textcolor{red}{\textbf{Red}}: best, \textcolor{blue}{blue}: second best.}
\label{tab:ood_comparison}
\resizebox{\textwidth}{!}{%
\setlength{\tabcolsep}{1pt}
\begin{tabular}{l ccc cccc cccc}
\toprule
\multirow{3}{*}{\makecell{Method}} & \multicolumn{7}{c}{\textbf{Revisiting}} & \multicolumn{4}{c}{\textbf{Exploration}} \\
\cmidrule(lr){2-8} \cmidrule(lr){9-12}
& \multicolumn{3}{c}{{long-term consistency}} & \multicolumn{4}{c}{{video quality}} & \multicolumn{4}{c}{{video quality}} \\
\cmidrule(lr){2-4} \cmidrule(lr){5-8} \cmidrule(lr){9-12}
& \makecell{PSNR$\uparrow$} & \makecell{SSIM$\uparrow$} & \makecell{LPIPS$\downarrow$} & \makecell{AQ(\%)$\uparrow$} & \makecell{IQ(\%)$\uparrow$} & \makecell{MS(\%)$\uparrow$} & \makecell{DD(\%)$\uparrow$} & \makecell{AQ(\%)$\uparrow$} & \makecell{IQ(\%)$\uparrow$} & \makecell{MS(\%)$\uparrow$} & \makecell{DD(\%)$\uparrow$} \\
\midrule

AC3D\cite{bahmani2025ac3d}
& 13.74 & 0.48 & 0.49
& 55.94 & 70.40 & 98.62
& \textcolor{first}{\textbf{97.46}}
& 57.20 & 73.25
& \textcolor{second}{99.11}
& \textcolor{second}{97.58} \\

DFoT\cite{song2025history}
& 14.80 & 0.47 & 0.32
& 47.44 & 71.30 & 98.34 & 29.29
& 47.76 & 72.04 & 98.54 & 24.40 \\

SEVA\cite{zhou2025stable}
& 12.31 & 0.48 & 0.45
& 50.10 & 73.30 & 98.50 & 90.44
& 54.18 & 73.96 & 98.44 & 97.36 \\

VMem\cite{li2025vmem}
& 15.43 & 0.49 & 0.27
& 54.95 & 69.12 & 97.45 & 60.98
& 55.69 & 70.11 & 97.97 & 60.98 \\

WorldPlay\cite{sun2025worldplay}
& 16.54 & 0.71 & 0.27
& \textcolor{second}{58.99}
& \textcolor{first}{\textbf{75.81}}
& 98.47 & 92.68
& 57.71 & 73.81 & 98.76 & 90.46 \\

WorldPlay-d\cite{sun2025worldplay}
& 16.57 & 0.69 & 0.26
& 58.44 & 75.70 & 98.81 & 93.00
& 57.69 & 74.22 & 98.94 & 95.35 \\

\midrule

Ours-Cog
& \textcolor{second}{20.79}
& \textcolor{second}{0.72}
& \textcolor{second}{0.23}
& 57.96
& 74.38
& \textcolor{first}{\textbf{98.91}}
& \textcolor{second}{95.00}
& \textcolor{second}{57.89}
& \textcolor{second}{75.83}
& \textcolor{first}{\textbf{99.12}}
& 97.50 \\

Ours-Wan
& \textcolor{first}{\textbf{21.35}}
& \textcolor{first}{\textbf{0.73}}
& \textcolor{first}{\textbf{0.22}}
& \textcolor{first}{\textbf{60.09}}
& \textcolor{second}{75.77}
& \textcolor{second}{98.84}
& 94.50
& \textcolor{first}{\textbf{59.83}}
& \textcolor{first}{\textbf{78.29}}
& 98.78
& \textcolor{first}{\textbf{97.70}} \\

\bottomrule
\end{tabular}%
}
\vspace{-5mm}
\end{table*}
\noindent\textbf{Computational and Data Efficiency.} 
\label{subsec:Efficiency}
Table~\ref{tab:efficiency} compares the extra computational overhead, extra parameters, and training data for methods with memory mechanisms.
Although WorldPlay~\cite{sun2025worldplay} introduces no additional parameters, it results in high computational costs due to the quadratic complexity of self-attention on the extended sequence. 
As shown in Table~\ref{tab:efficiency}, the memory operations in WorldPlay~\cite{sun2025worldplay} require {3,065 TFLOPs}, which is approximately $3.3\times$ its backbone (923.03 TFLOPs) and accounts for the majority of the total inference cost.
Our method adopts a separate, lightweight branch for memory modeling. 
This design adds {128.42M} parameters to the CogVideoX~\cite{yang2024cogvideox} backbone (about 2.2\% of the total), but significantly reduces computational load. 
Specifically, the memory branch in our method requires only {2.97 TFLOPs}, which is over $1,000\times$ lower than that of WorldPlay~\cite{sun2025worldplay}. 
As a result, the total inference cost of our method (443.78 TFLOPs) is close to the backbone-only cost. {More Comparisons on memory and latency are provided in the \textbf{Appendix}.}

In terms of training data efficiency, our method needs only \textbf{14K} training samples, while WorldPlay~\cite{sun2025worldplay} requires \textbf{320K} samples, about $23\times$ higher than ours. Overall, our approach achieves effective memory modeling with substantially lower computational and data costs than prior works.  

\begin{table*}[t]
	\centering
	\caption{Quantitative comparison across three challenging revisiting trajectories. IQ: Imaging Quality. \textcolor{red}{\textbf{Red (bold)}}: best; \textcolor{blue}{blue}: second best.}
	\label{tab:comparison}
	\resizebox{\textwidth}{!}{%
	\setlength{\tabcolsep}{4pt}
	\begin{tabular}{l ccc ccc ccc}
	\toprule
	\multirow{2}{*}{\makecell{Method}} & \multicolumn{3}{c}{\textbf{Repeated Revisiting}} & \multicolumn{3}{c}{\textbf{Random Loop Insertion}} & \multicolumn{3}{c}{\textbf{Spatially-Offset Return}} \\
	\cmidrule(lr){2-4} \cmidrule(lr){5-7} \cmidrule(lr){8-10}
	& PSNR$\uparrow$ & LPIPS$\downarrow$ & IQ(\%)$\uparrow$ & PSNR$\uparrow$ & LPIPS$\downarrow$ & IQ(\%)$\uparrow$ & PSNR$\uparrow$ & LPIPS$\downarrow$ & IQ(\%)$\uparrow$ \\
	\midrule
	AC3D~\cite{bahmani2025ac3d} 
	& 13.16 & 0.47 & 55.09
	& 16.49 & 0.32 & 63.58
	& 10.50 & 0.56 & 65.38 \\
	DFoT\cite{song2025history}
	& 14.39 & 0.39 & 66.44 
	& \textcolor{blue}{20.62} & \textcolor{red}{\textbf{0.13}} & 70.39 
	& 10.41 & 0.55 & 67.24 \\
	SEVA\cite{zhou2025stable}
	& 13.26 & 0.51 & 53.53 
	& 18.99 & 0.20 & \textcolor{red}{\textbf{75.38}} 
	& 9.71 & 0.71 & 68.04 \\
	VMem\cite{li2025vmem}
	& 18.91 & 0.30 & 59.71 
	& 19.03 & 0.25 & 65.50 
	& \textcolor{red}{\textbf{27.34}} & \textcolor{red}{\textbf{0.15}} & 61.87 \\
	WorldPlay~\cite{sun2025worldplay} 
	& 19.24 & 0.19 & \textcolor{blue}{73.17} 
	& 17.40 & 0.30 & 72.23 
	& 13.95 & 0.48 & \textcolor{blue}{68.96} \\
	WorldPlay-d~\cite{sun2025worldplay} 
	& 17.09 & 0.26 & \textcolor{red}{\textbf{73.85}} 
	& 17.61 & 0.53 & 71.98 
	& 11.79 & 0.70 & 67.72 \\
	\midrule
	Ours-Cog
	& \textcolor{blue}{21.92} & \textcolor{blue}{0.15} & 59.38 
	& 19.53 & 0.22 & 63.88 
	& 14.35 & 0.44 & 64.69 \\
	Ours-Wan 
	& \textcolor{red}{\textbf{22.03}} & \textcolor{red}{\textbf{0.14}} & 71.48 
	& \textcolor{red}{\textbf{21.76}} & \textcolor{blue}{0.16} & \textcolor{blue}{72.33} 
	& \textcolor{blue}{16.38} & \textcolor{blue}{0.29} & \textcolor{red}{\textbf{73.27}} \\
	\bottomrule
	\end{tabular}%
	}
	\vspace{-3mm}
\end{table*}

\subsection{Robustness Analysis}
\label{sec:additional_exp}

To further evaluate the robustness and adaptability of our decoupled framework, we conduct additional experiments on out-of-distribution (OOD) data and more challenging camera trajectories.

\noindent\textbf{OOD Generalization.}
{Table~\ref{tab:ood_comparison} evaluates the generalization on unseen distributions. Our methods achieve the top two results across all long-term consistency metrics while maintaining
competitive video quality in both Revisiting and Exploration. It also achieves the second-best IQ in Revisiting and leading IQ and DD in Exploration. Compared to WorldPlay~\cite{sun2025worldplay}, our method shows a significant leap (+4.81 PSNR, -0.05 LPIPS), validating that our decoupled memory mechanism enables robust generalization to novel domains and styles. We provide more visual analysis in the \textbf{Appendix}.}

\noindent\textbf{Complex Trajectories.}
As discussed in Sec.~\ref{sec:challenging_traj}, we evaluate performance under three challenging trajectory settings (Table~\ref{tab:comparison}), where our models continue to excel in both generation quality and spatial consistency.
Specifically, Ours-Wan achieves state-of-the-art consistency in \textit{Repeated Revisiting} (22.03 PSNR), and excels in \textit{Random Loop Insertion} with 21.76 PSNR.
For the \textit{Spatially-Offset Return} path, we evaluate consistency by computing metrics between the first and last frames. Although VMem~\cite{li2025vmem} achieves strong first-last frame consistency, it struggles to maintain high visual quality throughout the long-range generation process. Our method achieves a balance between video generation quality and spatial consistency, attaining 16.38 PSNR while maintaining an IQ of 73.27\%.

\begin{table}[t]
\centering
\caption{Ablation studies on hybird memory representation.}
\label{tab:ablation}
\footnotesize
\begin{tabular}{l|ccccc}
\toprule
Method & PSNR$\uparrow$ & SSIM$\uparrow$ & LPIPS$\downarrow$ & IQ$\uparrow$ & MS$\uparrow$ \\
\midrule
Only $K^t$ & 16.93 & 0.59 & 0.29 & 73.13\% & 98.98\% \\
Only $K^s$ & 20.78 & 0.69 & 0.23 & 72.18\% & 97.55\% \\
\textbf{Ours} & 20.79 & 0.72 & 0.23 & 74.38\% & 98.91\% \\
\bottomrule
\end{tabular}
\vspace{-3mm}
\end{table}

\begin{figure}[t]
\centering
\includegraphics[width=\linewidth]{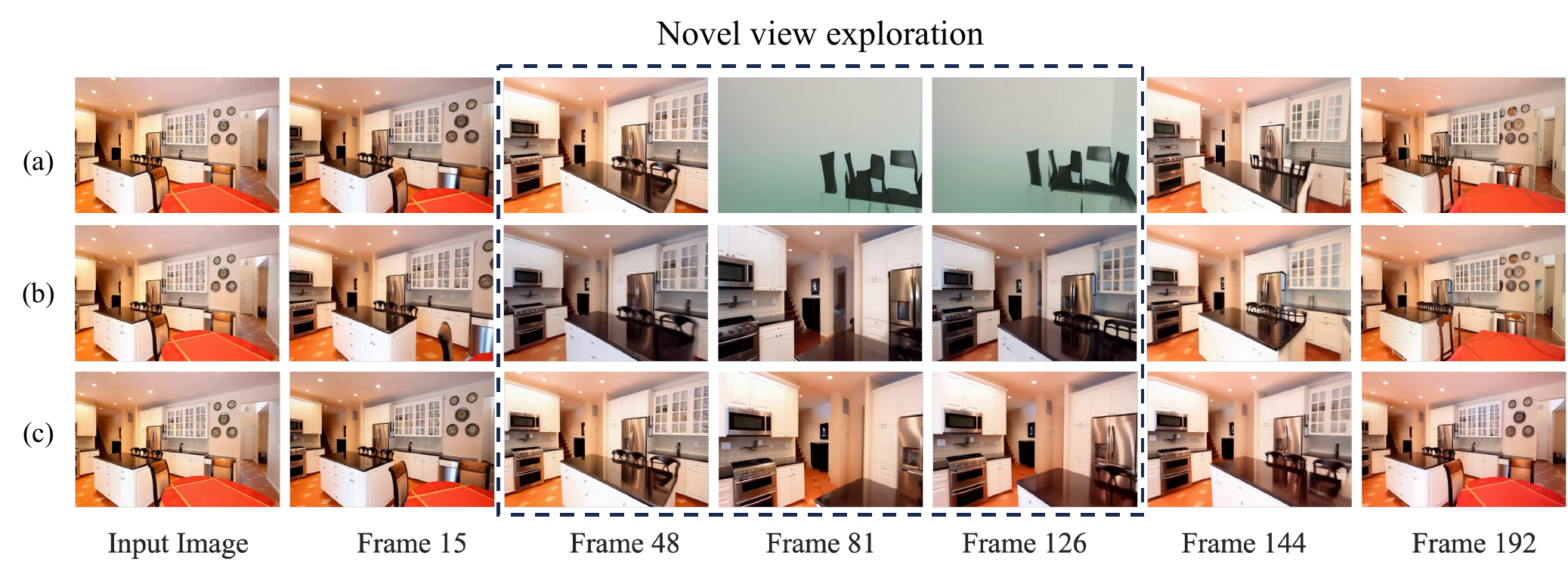}
\vspace{-5mm}
\caption{
\textbf{Ablation Study} on camera-aware gating mechanism and history dropout strategies.
(a) Baseline without gating fails on novel viewpoints.
(b) Adding the camera-aware gate improves realism in novel views but causes inconsistency.
(c) Full model with history dropout achieves both realistic and temporally consistent results.
}
\label{fig:ablation_gate}
\vspace{-2mm}
\end{figure}

{\subsection{Ablation Studies}}
\label{sec:ablation_studies}

We conduct ablation studies using our CogVideoX-based variant, i.e., Ours-Cog, selected for its computational efficiency and smaller parameter count.

\noindent\textbf{Hybrid Memory Representation.} 
Table~\ref{tab:ablation} presents the ablation study on the two components of our hybrid memory representation using out-of-distribution (OOD) data. 
When using only the temporal memory representation ($K^t$), the generated videos achieve the highest MS score of 98.98\%, outperforming the spatial-only variant ($K^s$, 97.55\%). 
This suggests that $K^t$ is particularly effective at capturing motion consistency across frames. 
In contrast, compared with the temporal-only variant, the ``Only $K^s$'' configuration substantially improves the spatial quality metrics, increasing PSNR from 16.93 to 20.78 and SSIM from 0.59 to 0.69, while reducing LPIPS from 0.29 to 0.23. It also achieves a slightly higher IQ score of 72.18\%. 
Combining $K^t$ and $K^s$ achieves the best overall performance, with the highest PSNR, SSIM, and IQ, confirming their complementary benefits for temporal coherence and spatial fidelity.

\noindent\textbf{Camera-aware Gating Mechanism.}
We evaluate the camera-aware gating mechanism and history dropout strategy in Figure~\ref{fig:ablation_gate}. The test sequence comprises a historical reference (Frame 15), a novel view exploration phase (Frames 48--126), and a revisiting phase (Frames 144--192).
The baseline without gating (Figure~\ref{fig:ablation_gate} (a)) maintains consistency during revisiting but suffers from texture artifacts in novel views.
Incorporating the camera-aware gate (Figure~\ref{fig:ablation_gate} (b)) effectively filters irrelevant history during exploration, significantly improving texture realism. However, this selective injection leads to temporal inconsistencies between the exploration and revisiting phases.
Finally, our history dropout strategy remedies this by enforcing robustness against missing historical context. As shown in Figure~\ref{fig:ablation_gate} (c), the full model achieves both high-fidelity synthesis for novel views and temporal coherence for revisited regions.

\section{Conclusion}
\vspace{-1mm}
We propose a decoupled memory-conditioned framework for spatially consistent long-horizon video generation. By separating memory modeling from the generation backbone, our method improves scene-revisit consistency while preserving generative capacity for novel regions with reduced training cost. We introduce a lightweight memory branch with hybrid temporal-spatial memory representation, per-frame cross-attention for spatially relevant conditioning, and camera-aware gating to activate memory only when valid historical references exist. Experiments show that our approach achieves state-of-the-art visual quality and spatial consistency across diverse camera trajectories while significantly reducing training data requirements.

\noindent\textbf{Limitations}. 
Following prior work, our method relies on FOV-based memory retrieval, where both the accuracy of FOV matching and camera pose control can affect the consistency and fidelity of generated revisited scenes. Moreover, as our framework builds upon existing video generation models, our performance upper bound is also constrained by the capability of the underlying base model.

\clearpage
{\small
\bibliographystyle{ieee_fullname}
\bibliography{Ref}
}

\newpage
\appendix

\setcounter{figure}{0}
\setcounter{table}{0}
\setcounter{equation}{0}

\renewcommand{\thefigure}{A\arabic{figure}}
\renewcommand{\thetable}{A\arabic{table}}
\renewcommand{\theequation}{A\arabic{equation}}

\section*{Appendix}

In this appendix, we provide the following materials:
\begin{itemize}
\item Section~\ref{sec:A}: The camera-aware gating algorithm (referring to Sec.~\ref{sec:gating} in the main paper);
{\item Section~\ref{sec:B}: More training and efficiency details, including training settings, FLOPs calculation, GPU memory, and wall-clock latency (referring to Sec.~\ref{sec:exp_setup}, Sec.~\ref{subsec:Efficiency} and Table~\ref{tab:efficiency} in the main paper);}
\item Section~\ref{sec:C}: More qualitative results (referring to Sec.~\ref{sec:additional_exp} in the main paper).
\item Section~\ref{sec:D}: Broader impact of our work.
\end{itemize}

\section{Algorithm of Camera-aware Gating}
\label{sec:A}

This section presents in detail our camera-aware gating mechanism, including the computation of the geometric relevance score $s_i$ and the filtering of unmatched frames based on spatial and temporal constraints. The goal is to determine when to enable memory interaction. We first enable all the gates and then deactivate the interaction in three steps, as described in the following.

\vspace{+2mm}
\noindent\textbf{Step 1: Deactivate Gate by Geometric Relevance Scoring} 

\noindent For each target pose, we calculate its geometric alignment with historical poses by computing the field-of-view (FOV) overlap ~\cite{xiao2025worldmem} and a normalized translation-distance penalty to derive a relevance score for each historical frame. 
The frame with the highest score is selected as the initial candidate and this score is set as the geometric relevance score $s_i$ for the target frame. A high $s_i$ suggests strong alignment with known content, while a low $s_i$ suggests exploration of new regions. 
If $s_i$ is below a threshold, we deactivate the corresponding gate due to insufficient alignment with historical memory.

\vspace{+2mm}
\noindent\textbf{Step 2: Deactivate Gate by Translation Distance} 

\noindent
We also deactivate the gate for candidates with large translation distances, even if their relevance scores are high. This step is crucial to avoid overestimating $s_i$ in scenarios such as forward-moving trajectories, where significant FOV overlap with earlier frames might occur. Introducing a  translation distance constraint helps prevent such false positives and enhances gating accuracy.

\vspace{+2mm}
\noindent\textbf{Step 3: Deactivate Gate due to Temporal Redundancy} 

\noindent
To prevent over-constraining the generation process, we apply temporal redundancy filtering across adjacent target frames. Injecting highly similar historical contexts into consecutive frames can weaken the temporal dynamics. In general, if a frame within a brief interval is grounded on historical memory, its neighboring frames will maintain spatial consistency. This filtering reduces the frequency of memory usage, balancing spatial coherence with temporal dynamics.

We summarize the whole algorithm of camera-aware gating in Algorithm~\ref{alg:reference_retrieval}. Frames with high $s_i$ (which remain active after the filtering steps) are treated as revisits to known regions, thus enabling memory injection, while frames with low $s_i$ (or those deactivated by distance/temporal constraints) are considered novel and generated without memory interaction.

\begin{algorithm}[t]
    \footnotesize
    \caption{Camera-Aware Gating via Geometric Relevance Scoring}
    \label{alg:reference_retrieval}
    \vspace{0.02cm}
    
    \textbf{Input:}
    \vspace{-0.25cm}
    \begin{itemize}\itemsep0pt \parskip0pt \parsep0pt \topsep2pt
        \item[$\bullet$] Target poses $P_{\mathrm{tgt}}=\{P_t\}_{t=1}^{N}$, where $P_t=(R_t,\mathbf{t}_t)\in \mathrm{SE}(3)$.
        \item[$\bullet$] Historical poses $P_{\mathrm{hist}}=\{P_r\}_{r=1}^{F}$, where $P_r=(R_r,\mathbf{t}_r)\in \mathrm{SE}(3)$.
    \end{itemize}
    \vspace{-0.2cm}
    
    \textbf{Output:}
    \vspace{-0.25cm}
    \begin{itemize}\itemsep0pt \parskip0pt \parsep0pt \topsep2pt
        \item[$\bullet$] Geometric relevance scores $S=\{s_t\}_{t=1}^{N}$.
        \item[$\bullet$] Binary gates $G=\{g_t\}_{t=1}^{N}$, where $g_t\in\{0,1\}$.
    \end{itemize}
    \vspace{-0.3cm}
    
    \textbf{begin}
    
    \noindent
    \hspace{0.6em}%
    \vrule width 0.3pt\hspace{0.7em}%
    \begin{minipage}{0.94\linewidth}
    \setlength{\abovedisplayskip}{3pt}
    \setlength{\belowdisplayskip}{3pt}
    \setlength{\abovedisplayshortskip}{2pt}
    \setlength{\belowdisplayshortskip}{2pt}
    
    \vspace{0.2cm}
    
    For each target pose index $t=1,\dots,N$:
    
    \noindent
    \hspace*{1em}%
    \vrule width 0.3pt\hspace{0.7em}%
    \begin{minipage}{0.90\linewidth}
    
    Initialize current gate:
    \[
    g_t \leftarrow 1
    \]
    
    Compute FOV overlap with all historical poses~\cite{xiao2025worldmem}:
    \[
    \rho^{(t)}_r \leftarrow \text{OverlapComputation}(P_t, P_r), \qquad r=1,\dots,F
    \]
    
    Compute normalized translation distances:
    \[
    d^{(t)}_r \leftarrow \operatorname{Norm}\!\left(\|\mathbf{t}_t-\mathbf{t}_r\|_2\right), \qquad r=1,\dots,F
    \]
    
    Compute geometric relevance score:
    \[
    c^{(t)}_r \leftarrow \rho^{(t)}_r - \lambda_d d^{(t)}_r, \qquad r=1,\dots,F
    \]
    
    Select the highest-scoring historical pose:
    \[
    r_t^{*} \leftarrow \arg\max_r c^{(t)}_r,
    \qquad
    s_t \leftarrow c^{(t)}_{r_t^{*}}
    \]
    
    $\triangle$ \textbf{Step 1: Deactivate Gate by Geometric Relevance Scoring}
    \[
    \text{if } s_t < \tau_c,
    \qquad
    g_t \leftarrow 0
    \]
    
    $\triangle$ \textbf{Step 2: Deactivate Gate by Translation Distance}
    
    \vspace{-0.3cm}
    \[
    \text{if } d^{(t)}_{r_t^{*}} > \tau_d,
    \qquad
    g_t \leftarrow 0
    \]
    
    \end{minipage}
    
    \vspace{0.1cm}
    For each target pose index $t=2,\dots,N$:
    
    \noindent
    \hspace*{1em}%
    \vrule width 0.3pt\hspace{0.7em}%
    \begin{minipage}{0.90\linewidth}
    
    \vspace{0.1cm}
    $\triangle$ \textbf{Step 3: Deactivate Gate due to Temporal Redundancy}
    
    \vspace{-0.3cm}
    \[
    \text{if } g_t=1 \land g_{t-1}=1 \land |r_t^{*}-r_{t-1}^{*}| < \tau_{\text{temp}},
    \qquad
    g_t \leftarrow 0
    \]
    \end{minipage}
    
    \vspace{0.1cm}
    \textbf{Return $S$ and $G$.}
    \vspace{0.10cm}
    \end{minipage}
    \vspace{0.1cm}
    \textbf{end}
    \end{algorithm}

\section{Experimental Details}
\label{sec:B}
\subsection{Training Details}
Our training follows a two-stage procedure. 
First, we train the camera control module following AC3D~\cite{bahmani2025ac3d}. 
Specifically, camera pose conditioning is injected only at the low-frequency denoising steps, where global structure is determined.
We observe that restricting camera supervision to these early denoising steps accelerates convergence and leads to more accurate camera pose control.
In the second stage, we freeze both the video backbone and the camera control module, and train the memory control branch independently. 
During the first $20\%$ of training iterations in this stage, the camera control module is temporarily disabled and the memory branch is trained in isolation, allowing it to first learn stable memory representations before interacting with the camera control signal.

\subsection{FLOPs Estimation Details}

We detail the FLOPs estimation protocol used in Table 2 of the main paper. 
For fair comparison, all methods are evaluated under the same setting: a single denoising step with 61 input frames. 
Backbone FLOPs refer to the computation cost of the pretrained generation model itself, while memory FLOPs refer to the additional cost introduced by the memory mechanism. 
The total FLOPs are reported as the sum of backbone FLOPs and memory FLOPs.

For VMEM \cite{li2025vmem}, the backbone FLOPs are calculated by running SEVA \cite{zhou2025stable} (historical frames=1) to generate 61 frames under its iterative inference paradigm. 
The memory FLOPs refer to the extra cost arising from its memory mechanism, which retains more historical frames(=4) and reduces the newly generated frames per iteration, thus increasing the total number of iterations required.

For WorldPlay\cite{sun2025worldplay}, Ours-Wan, and Ours-Cog, the backbone FLOPs are calculated by running the pretrained video generation models (HunyuanVideo~1.5~\cite{kong2024hunyuanvideo}, Wan~2.1~\cite{wan2025wan}, and CogVideoX~\cite{yang2024cogvideox}), respectively, to generate 61 frames with camera pose conditioning. 
For WorldPlay\cite{sun2025worldplay}, the memory FLOPs arise from concatenating retrieved memory frames into the input sequence, which increases quadratically the attention cost of the backbone. 
For our methods, the memory FLOPs arise from the computation of the lightweight memory branch, which operates separately from the backbone and introduces negligible overhead. 

{\subsection{ Peak Memory and Wall-Clock Latency}
We also report the peak GPU memory and wall-clock latency on an NVIDIA A800 GPU in Table\ref{tab:memory_latency}. The results distinguish the backbone-only cost from the total cost with memory modeling, directly showing the additional practical overhead.
\begin{table}[t]
\centering
\caption{Peak GPU memory and wall-clock latency under the same inference setting. 
Latency is measured per denoising step.}
\label{tab:memory_latency}
\resizebox{\linewidth}{!}{
\begin{tabular}{lcccccc}
\toprule
\multirow{2}{*}{Method}
& \multicolumn{3}{c}{Peak Memory (GB)}
& \multicolumn{3}{c}{Wall-Clock Latency (s)} \\
\cmidrule(lr){2-4}
\cmidrule(lr){5-7}
& Backbone & Total & Increase
& Backbone & Total & Increase \\
\midrule
WorldPlay
& 21.57 & 27.03 & +5.46 (+25.3\%)
& 11.18 & 45.47 & +34.29 (+306.7\%) \\

Ours-Cog
& 26.02 & 31.26 & +5.24 (+20.1\%)
& 5.78 & 5.97 & +0.20 (+3.5\%) \\

Ours-Wan
& 41.30 & 48.89 & +7.60 (+18.4\%)
& 9.29 & 10.81 & +1.52 (+16.4\%) \\
\bottomrule
\end{tabular}
}
\end{table}

}
\section{More Qualitative Results}
\label{sec:C}

\begin{figure*}[t]
	\centering
	\includegraphics[width=\linewidth]{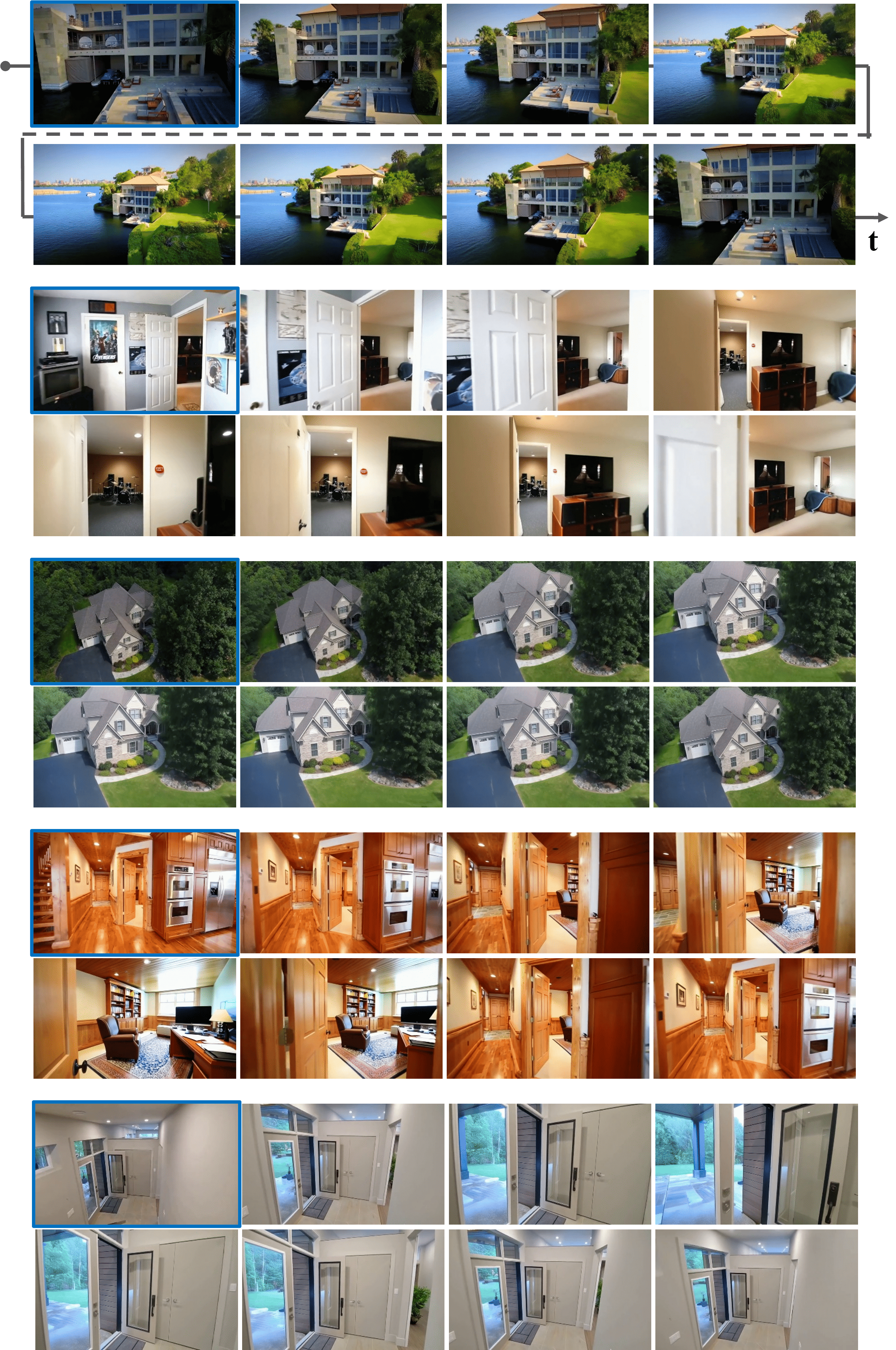}
    \vspace{-5mm}
\caption{\textbf{Qualitative results on the RealEstate10K dataset~\cite{Stereo}.}  The \textcolor{blue}{boxed} frame marks the starting frame, with following sequence from the first row to the second.} 
	\label{fig:supp_realestate}. 
\vspace{-5mm}
\end{figure*}

\begin{figure*}[t]
	\centering
	\includegraphics[width=\linewidth]{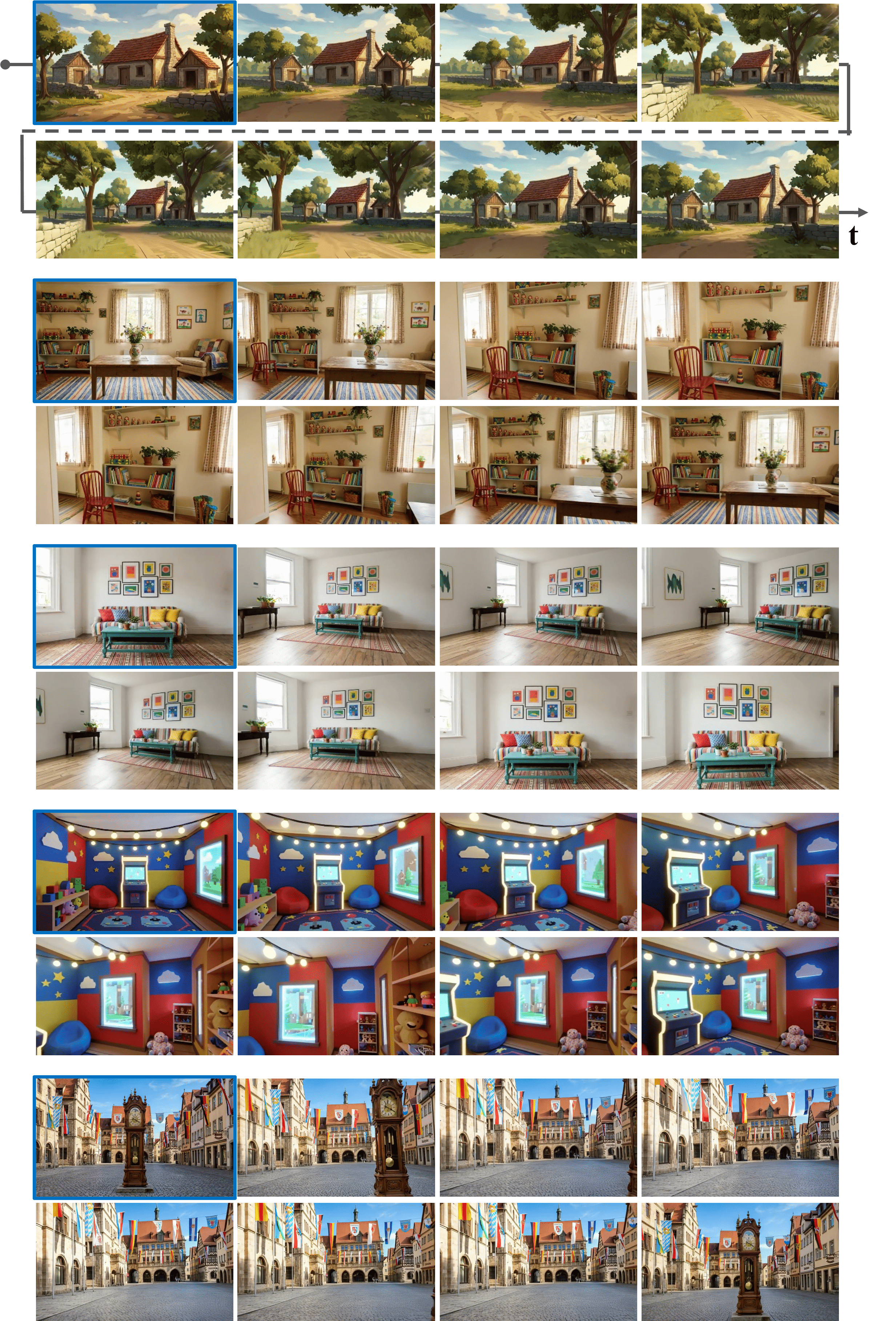}
    \vspace{-5mm}
\caption{\textbf{Qualitative results on out-of-distribution (OOD) scenes}, with diverse styles in both indoor and outdoor environments.
The \textcolor{blue}{boxed} frame marks the starting frame, with the following sequence from the first row to the second.} 
	\label{fig:supp_ood}
\vspace{-5mm}
\end{figure*}

\begin{figure*}[t]
	\centering
	\includegraphics[width=0.90\linewidth]{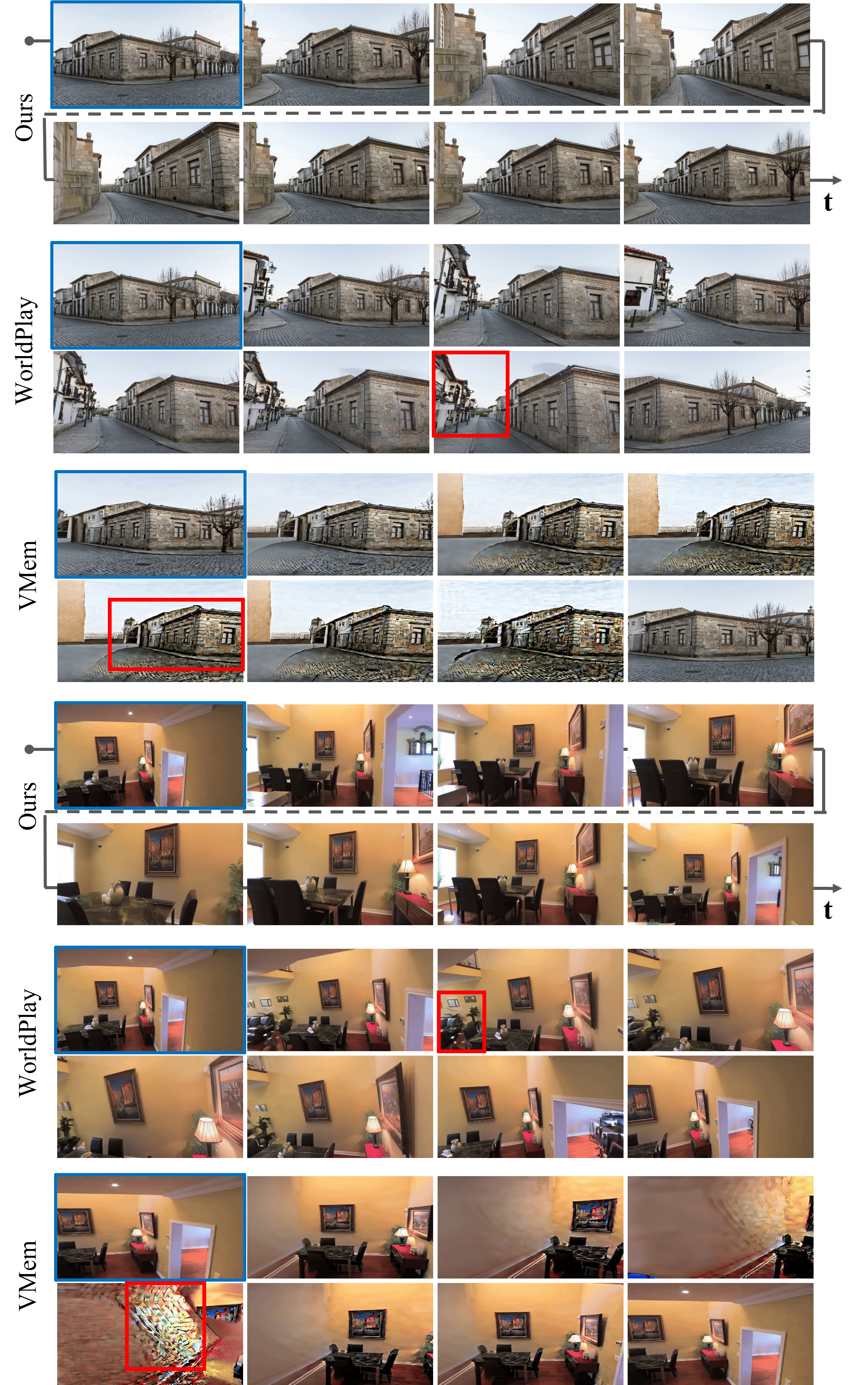}
\caption{\textbf{Comparison with memory-based baselines.}
Top: an OOD example. Bottom: a  RealEstate10K~\cite{Stereo} example. 
\textcolor{blue}{boxed} frame marks the starting frame, with the following sequence from the first row to the second.
\textcolor{red}{Red} boxes mark artifacts of VMem~\cite{li2025vmem} and WorldPlay~\cite{sun2025worldplay} when generating novel contents in unseen regions. }
	\label{fig:supp_compare}
\end{figure*}

\noindent\textbf{More results on RealEstate10K and OOD scenes.}
We provide more qualitative examples on RealEstate10K~\cite{Stereo} in Figure~\ref{fig:supp_realestate} and additional OOD results in Figure~\ref{fig:supp_ood}, including scenes with various artistic styles in indoor and outdoor environments. 
Our method not only synthesizes reasonable novel contents during scene exploration, but also preserves scene consistency when the camera revisits previously observed views. 
Although trained only on RealEstate10K, our decoupled design preserves the generative capability of the pretrained video model, enabling strong generalization to diverse scenes beyond the training distribution.

\noindent\textbf{Additional comparisons with memory-based baselines.}
Figure~\ref{fig:supp_compare} provides two additional comparison examples with memory-based baselines, including VMem~\cite{li2025vmem} and WorldPlay~\cite{sun2025worldplay}, on OOD data and RealEstate10K. 
As highlighted by the red boxes, prior methods tend to introduce visible artifacts when synthesizing novel contents in unseen regions. 
In contrast, our method generates cleaner and more coherent results, demonstrating superior robustness in scene exploration and revisiting.
\section{Broader Impact}
\label{sec:D}
This work improves spatial consistency and visual realism during scene revisiting in long-horizon video generation, and can be applied to embodied intelligence, world simulation, game content generation, and virtual environment construction.

However, more coherent and realistic long-video generation capabilities may also be misused to create misleading synthetic content, such as fabricated scene walkthroughs or surveillance-style videos. Since the proposed method can reduce artifacts such as spatial inconsistency, the generated content may become more difficult to identify as synthetic videos.

Potential mitigation measures include synthetic-content watermarking, provenance tracking, sensitive-data filtering, controlled release of model weights, and clear labeling of generated content. This work is primarily intended as foundational research but not direct deployment in high-risk scenarios.

\clearpage

\end{document}